%% file: revision.tex
\documentclass[10pt,journal,compsoc]{IEEEtran}

\ifCLASSOPTIONcompsoc
  \usepackage[nocompress]{cite}
\else
  \usepackage{cite}
\fi

\ifCLASSINFOpdf

\else

\fi

\usepackage[utf8]{inputenc} 
\usepackage[T1]{fontenc}    
\usepackage{url}            
\usepackage{booktabs}       
\usepackage{nicefrac}       
\usepackage{microtype}      
\usepackage{amsthm}
\usepackage{amsbsy}
\usepackage{amsfonts}
\usepackage{amssymb}
\usepackage{stmaryrd}
\usepackage{xcolor}
\usepackage{graphicx}
\usepackage{amsmath,mathtools}
\usepackage{algorithm,algorithmic}
\usepackage{subfigure}
\usepackage{mathtools}

\begin{document}
\input{header}
%
\title{TAKDE: Temporal Adaptive Kernel Density Estimator for Real-Time Dynamic Density Estimation}

\author{Yinsong Wang, Yu Ding,~\IEEEmembership{Senior Member,~IEEE,}
        and~Shahin~Shahrampour,~\IEEEmembership{Senior Member,~IEEE}
\IEEEcompsocitemizethanks{\IEEEcompsocthanksitem Y. Wang and S. Shahrampour are with the Department of Mechanical and Industrial Engineering at Northeastern University, Boston, MA 02115 USA.\protect\\
E-mail: {\tt\small wang.yinso@northeastern.edu}\\ E-mail: {\tt\small s.shahrampour@northeastern.edu}
\IEEEcompsocthanksitem Y. Ding is with the Wm Michael Barnes '64 Department of Industrial and Systems Engineering at Texas A\&M University, College Station, TX 77843 USA.\\
E-mail: {\tt\small yuding@tamu.edu}}}



\IEEEtitleabstractindextext{%
\begin{abstract}
Real-time density estimation is ubiquitous in many applications, including computer vision and signal processing. Kernel density estimation is arguably one of the most commonly used density estimation techniques, and the use of ``sliding window'' mechanism adapts kernel density estimators to dynamic processes. In this paper, we derive the asymptotic mean integrated squared error (AMISE) upper bound for the ``sliding window'' kernel density estimator. This upper bound provides a principled guide to devise a novel estimator, which we name the temporal adaptive kernel density estimator (TAKDE).  Compared to heuristic approaches for ``sliding window'' kernel density estimator, TAKDE is theoretically optimal in terms of the worst-case AMISE. We provide numerical experiments using synthetic and real-world datasets, showing that TAKDE outperforms other state-of-the-art dynamic density estimators (including those outside of kernel family). In particular, TAKDE achieves a superior test log-likelihood with a smaller run-time. 
\end{abstract}

}

\maketitle

\IEEEdisplaynontitleabstractindextext

\IEEEpeerreviewmaketitle

\IEEEraisesectionheading{\section{Introduction}\label{sec:introduction}}

\IEEEPARstart{T}his work is concerned with estimation and tracking of dynamic probability density functions in real time, motivated by a nanoscience application. The introduction of \emph{in situ} transmission electron microscope (TEM) technology \cite{zheng2009observation} allows the growth of nanoparticles to be captured in real time and has the potential to enable precise control in nanoparticle self-assembly processes.  Part of the underlying nanoscience problem is framed into a learning problem with the following characteristics \cite{qian2019fast}: (1) Estimation and tracking of a time-varying probability density function that reflects the collective changes across ensembles of the nano objects. (2) It seems inevitable to adopt a non-parametric approach in the density tracking, because there is no settled parametric density function that can adequately describe growth mechanisms in a multi-stage nanoparticle growth process \cite{woehl2014direct, zheng2009observation}. (3) In order to be useful for in-process decision making, the density estimation and tracking needs to be conducted in real time.  By "real-time" we mean that the learning and computation speed ought to be fast enough relative to the imaging rate (or the data arrival rate in general), which is $15$ frames per second (fps) in \cite{zheng2009observation}. While the research is motivated by the dynamic nano imaging, we believe that the aforementioned characteristics are rather common in many types of dynamic streaming data, brought forth in various applications by fast-pace data collection capability. The objective of this research is to present one competitive solution for dynamic density estimation and tracking.


On the subject of density estimation, kernel density estimator has had great success (in terms of accuracy) for static datasets \cite{scott2015multivariate}. The direct adaptation of kernel density estimator to dynamic density estimation \cite{kristan2010online} is infeasible as the memory and computation cost constantly scale with the total number of incoming data points. \cite{hang2018kernel} further shows that even with unlimited computation and storage resources, a traditional kernel density estimator will only be a consistent estimator for a few specific dynamic systems. \cite{qian2019fast} also shows that traditional kernel density estimation falls short in practice in dynamic density estimation due to limited data availability.

To address the disadvantages of traditional kernel density estimator in dynamic density estimation, most researchers resort to the "sliding window" mechanism \cite{zhou2003m,heinz2008cluster,qahtan2016kde}. For example, \cite{zhou2003m} proposed the M-kernel algorithm, where the contribution of each data point in the "sliding window" is approximated as an additional weight added to the kernel density at the closest grid point. This approach manages to keep the memory and computation costs within budget despite the growth of the total number of data points. However, with a poor choice of grid points, it can suffer from either over-fitting or under-fitting. \cite{heinz2008cluster} employed cluster kernel and resampling technique to improve the merger performance. This approach uses the exponentially decaying weight scheme to capture the dynamic 
of the true density. \cite{boedihardjo2008framework} proposed the local region kernel density estimator (LRKDE), where the kernel bandwidth varies in different regions. The regions are divided such that the sum of data variances in each region is minimized. LRKDE also uses a "sliding window" to capture the dynamic of the true density. \cite{qahtan2016kde} further improved upon the previous works by using linear interpolation with kernel densities at grid points to approximate the kernel density estimator and then updating the kernel densities at the grid points with data points within a "sliding window".

The "sliding window" kernel density estimators do not only use the data points at the current time stamp, and they take into account older data points for inferring the current distribution. Intuitively, this mechanism provides two improvements that allow the kernel density estimator to work well in dynamic density estimation. First, defining a window size according to the computation and memory limit of the learning machine can alleviate the scalability issue of the kernel density estimator as old data points that are irrelevant to the current distribution can be discarded. Second, including older data points in the window can help alleviate the low data volume issue for most streaming data applications. However, to the best of our knowledge, all "sliding window" kernel density estimators proposed so far focus on modifying the kernel density estimator itself, and less attention has been given to the "sliding window" mechanism. As the only component that addresses the "dynamic" part of dynamic density estimation, there is no answer regarding how this mechanism affects the performance of the estimation.

We note that there also exists another line of works that model the dynamic density transition using a dynamical system with a fixed number of parameters. One class of frameworks is based on Bayesian learning \cite{caron2007generalized,rodriguez2008bayesian,mena2016dynamic}, which models the prior with an evolving Dirichlet process called dependent Dirichlet process, where the dependence between a class of Dirichlet processes is defined by a covariate. When using the covariate to describe time, the dependent Dirichlet process can be used to model the evolution of the dynamic distribution. The computation and memory costs are also maintained at a constant level. Another approach \cite{qian2019fast} couples B-spline with Kalman filter to capture the density evolution with a state space model. It imposes space continuity with B-spline smoothing and time continuity with Kalman filter to develop a fast density estimator for real-time process control. However, these estimators always need a normalization process with numerical operations to return a proper density function. For real-time density estimation tasks that require a model update cycle in the order of sub-second, these methods may not be ideal as we will later show in simulations.

In this paper, we propose the temporal adaptive kernel density estimator (TAKDE), a novel kernel density estimator for real-time dynamic density estimation that is theoretically optimal in terms of the worst-case asymptotic mean integrated squared error (AMISE). For the first time, we derive the AMISE upper bound for the "sliding window" kernel density estimator in a dynamic density estimation context. The minimizer of the upper bound entails a novel sequence for bandwidth selection and data weighting, which forms the basis of TAKDE. We provide numerical experiments on synthetic datasets to support our theoretical claim, and we then use several real-world datasets to show that TAKDE outperforms other state-of-the-art fast dynamic density estimators, such as the B-spline Kalman Filter \cite{qian2019fast} and KDEtrack \cite{qahtan2016kde} in terms of mean test log-likelihood metric. Interestingly, TAKDE also dominates these algorithms in terms of achieving a smaller run-time.

The organization of the paper is as follows. We present in Section \ref{sec:preliminaries} the preliminaries, including definitions and notations used throughout the paper. In Section \ref{sec:algorithm}, we present the details for TAKDE design, which addresses three important questions, i.e., the selection of window size, bandwidth and the data weights. We provide in Section \ref{sec:experiment} numerical experiments with synthetic and real datasets to demonstrate the performance of TAKDE. Finally, we draw conclusions and discuss the potential and limitations of TAKDE in Section \ref{sec:conclusion}.

\section{Preliminaries}\label{sec:preliminaries}

\subsection{Kernel Density Estimation: A Brief Overview}
The kernel density estimator for a given set of data points $\{x_i\}_{i=1}^n$ is as follows
\begin{equation}\label{eq:kde}
    \hat{p}(x~;\sigma) = \frac{1}{n}\sum_{i=1}^{n}K_{\sigma}(x-x_i),
\end{equation}
where $K_{\sigma}(\cdot)$ is the kernel function with the bandwidth $\sigma$. Throughout this paper, $K(\cdot)$ denotes a standard kernel function with a unit kernel bandwidth. We have that $K_{\sigma}(x) = \frac{1}{\sigma}K(\frac{x}{\sigma})$. We further impose the following mild assumptions on the kernel function $K(\cdot)$.

\begin{assumption}\cite{wand1994kernel}\label{assump:kernel}
The bandwidth sequence $\sigma_n$ (the subscript n shows the dependence of $\sigma$  to the number of data points) has the following properties
    \begin{equation}
        \begin{aligned}\label{eq:band}
            \underset{n \rightarrow \infty}{\lim} \sigma_n &= 0\\
            \underset{n \rightarrow \infty}{\lim} n\sigma_n &= \infty,
        \end{aligned}
    \end{equation}
which implies that the bandwidth $\sigma_n$ decays slower than $n^{-1}$ and converges to $0$.
The standard kernel function $K(\cdot)$ is a bounded, symmetric probability density function with a zero first moment and a finite second moment. That is, the following properties hold
    \begin{equation}
        \begin{aligned}\label{properties}
            \int K(x)dx &= 1\\
            \int xK(x)dx &= 0\\
            \int x^2K(x)dx &< \infty.
        \end{aligned}
    \end{equation}
\end{assumption}
The convergence to $0$ for bandwidth is rather intuitive, in that when we have infinitely many data points at hand, our estimator can be as flexible as possible without having to be concerned about over-fitting. It is also easy to verify that many commonly used kernels (e.g., the Gaussian kernel $K(x) = \frac{1}{\sqrt{2\pi}}e^{-x^2}$) satisfy \eqref{properties}.

\subsection{Problem Formulation}
In dynamic estimation, the density evolves over time. The evolution might be continuous in nature, but 
we only observe samples from time to time. Here, we consider the case where the streaming data comes in batches. We first define the dynamic streaming dataset, where we observe one new batch of data points $\xb^{(t)}=\{x_i^{(t)} \in \R\}_{i=1}^{n_t}$ at a new time stamp $t$. This data structure applies to most real-world streaming datasets. An important example is estimating density information in video datasets \cite{qian2019fast} like the dynamic nano imaging problem mentioned in the introduction. An image processing tool extracts the sizes of nanoparticles as the sample points for estimating the normalized particle size distribution (NSPD), which is an indicator to anticipate and detect phase changes in the nanoparticle growth. This data structure further applies to many time-series datasets \cite{UCRArchive2018}. For the cases where streaming data comes in on a per point basis, one can convert those types of data into our defined structure through combining consecutive data points into batches.

We assume data points $\xb^{(t)}$ are generated independently from $p_t(x)$, the true density at time stamp $t$. Also, the data points $\xb^{(t)}$ and $\xb^{(t')}$ in different time stamps ($t\neq t'$) are independent from each other. We impose the following assumption on the true density function.

\begin{assumption}\label{assump:density}
The true density function $p_t(x)$ at any time stamp $t$ is twice differentiable, and its second derivative $p_t''(x)$ is continuous and square integrable.
\end{assumption}
Assumption \ref{assump:density} is commonly used for continuous density functions \cite{wand1994kernel}. {\rd The square integrable condition is necessary as the integrated second order Taylor expansion appears later in the error bound derivation.} 

Following \eqref{eq:kde}, we write the traditional kernel density estimator of the density $p_t(x)$ as follows
\begin{equation}\label{eq:kdet}
    \hat{p}_t(x~;\sigma) = \frac{1}{n_t}\sum_{i=1}^{n_t}K_{\sigma}(x-x^{(t)}_i).
\end{equation}

The "sliding window" kernel density estimator, popularly used in dynamic density estimation \cite{zhou2003m,heinz2008cluster,qahtan2016kde}, takes the following form
\begin{equation}\label{eq:swkde}
    \hat{h}_t(x) = \sum_{j\in \Tc_t} \alpha^{(t)}_j\hat{p}_j(x~;\sigma^{(t)}_j),
\end{equation}
where $\Tc_t$ represents the set of batches within the moving window (memory), $\hat{p}_j$ is defined following \eqref{eq:kdet}, and $\alpha^{(t)}_j$ is a non-negative weight sequence that satisfies $\sum_{j \in \Tc_t}\alpha^{(t)}_j = 1$, to ensure that the output is a proper density function. The window size is $T_t$, i.e. $|\Tc_t|=T_t$, so that $\Tc_t$ can be naturally written as $\Tc_t=\{t-T_t+1,\ldots,t\}$. The superscripts $(t)$ on $\alpha$ and $\sigma$ are omitted hereafter for the presentation clarity.

In order to develop a fast real-time estimator, we need to address the following three problems.
\begin{problem}\label{prob1}
How do we choose the set $\Tc_t$ to have a good enough "memory" for estimating the density at time $t$ {\rd while maintaining real-time processing}?
\end{problem}

\begin{problem}\label{prob2}
How do we design the weight sequence in \eqref{eq:swkde}?
\end{problem}

\begin{problem}\label{prob3}
How do we devise a kernel bandwidth selector in \eqref{eq:kdet}?
\end{problem}

\section{Algorithm Design}\label{sec:algorithm}

In this section, we derive the AMISE upper bound for the general "sliding window" kernel density estimator in \eqref{eq:swkde}. We then present a novel weight and bandwidth sequence, entailed by the upper bound minimizer (Problems \ref{prob2}-\ref{prob3}). We use these sequences to design the TAKDE algorithm.

\subsection{Asymptotic Mean Integrated Squared Error Upper Bound}
AMISE is a popular metric used to theoretically evaluate the performance of a density estimator \cite{wand1994kernel}. For a given density estimator $\hat{h}(x)$ of a density function $p(x)$, the mean integrated squared error (MISE) is defined as follows
\begin{equation}
\begin{aligned}
    MISE(\hat{h},p) &\triangleq \int \E[(\hat{h}(x)-p(x))^2] dx\\
    &= \int MSE(\hat{h},p) dx,
\end{aligned}
\end{equation}
where the expectation is taken with respect to the {\rd distributions of data points involved in estimator $\hat{h}$}. MISE is the integration of the mean squared error of the density estimator over the support. \cite{wand1994kernel} shows that the asymptotic expression (with respect to the sample size $n$) of the MISE for a standard kernel density estimator $\hat{p}(x~;\sigma_n)$ with kernel bandwidth $\sigma_n$ is
\begin{equation}\label{eq:mise}
    AMISE(\hat{p},p) = \frac{R(K)}{n\sigma_n} + \frac{1}{4}\sigma_n^4\mu_2^2(K)R(p''),
\end{equation}
where
\begin{equation}
\begin{aligned}
    R(f) &= \int f^2(x) dx,\\
    \mu_2(f) &= \int x^2f(x)dx.
\end{aligned}
\end{equation}
We can see that the conditions in \eqref{eq:band} guarantee that AMISE converges to zero as $n\rightarrow \infty$. {\rd The MISE and AMISE have been popular measures for characterizing non-parametric density estimators, including binned density estimator \cite{scott1985kernel}, kernel density estimator \cite{wand1994kernel}, wavelet density estimator \cite{hall1995formulae}, and diffusion estimator with a static limit \cite{botev2010kernel}. The exact expression for kernel density estimator can also be derived in the case of specific distributions like Gaussian distribution \cite{wand1994kernel}. However, all these derivations assume that data points in the non-parametric density estimator are samples from a static target density function.}

In the following theorem, we {\rd derive} the theoretical upper bound of AMISE for the "sliding window" kernel density estimator given in \eqref{eq:swkde} in the context of dynamic density estimation. {\rd To the best of our knowledge, this is the first AMISE bound for "sliding window" kernel density estimator in estimating the evolving true density $p_t(x)$.}

\begin{theorem}\label{the:upperbound}
Let Assumptions \ref{assump:kernel}-\ref{assump:density} hold. The AMISE of a "sliding window" kernel density estimator $\hat{h}_t$ at time $t$ with window size $|\Tc_t| = T_t$, weight sequence $\{\alpha_i\}_{i=1}^{T_t}$, and bandwidth sequence $\{\sigma_i\}_{i=1}^{T_t}$ has the following upper bound
\begin{equation}\label{eq:amise}
    \begin{aligned}
    AMISE(\hat{h}_t,p_t) \leq & \sum_{i \in \Tc_t} \frac{\alpha_i^2}{n_i\sigma_i}R(K)\\
    &+ (2T_t-1)\sum_{i \in \Tc_t}\alpha^2_iR(b^{(t)}_i) \\
    &+ \frac{2T_t-1}{4}\mu_2^2(K)\sum_{i \in \Tc_t}\alpha_i^2\sigma_i^4R(p_i''),
    \end{aligned}
\end{equation}
{\rd 
where $b_i^{(j)}(x)$ defines the difference between density functions $p_i(x),p_j(x)(j\geq i)$
\begin{equation}\label{eq:diff}
    b_i^{(j)}(x) \triangleq p_i(x) - p_j(x).
\end{equation}
}
\end{theorem}
\begin{proof}
We omit the superscript $(t)$ for weight $\alpha$ and bandwidth $\sigma$ for the presentation clarity. First, recall the definition of $\hat{h}_t(x)$ from \eqref{eq:kdet}-\eqref{eq:swkde}, where we have
\begin{equation}
    \hat{h}_t(x) = \sum_{i\in \Tc_t} \alpha_i\hat{p}_i(x~; \sigma_i)=\sum_{i \in \Tc_t} \frac{\alpha_i}{n_i}\sum_{j=1}^{n_i}K_{\sigma_i}(x-x^{(i)}_j).
\end{equation}
The bias of the estimator can be written as 
\begin{equation}
\begin{aligned}\label{bias}
    B(\hat{h}_t(x)) &\triangleq \E[\hat{h}_t(x)-p_t(x)]\\
    &=\E\Big[\sum_{i \in \Tc_t} \frac{\alpha_i}{n_i}\sum_{j=1}^{n_i}K_{\sigma_i}(x-x^{(i)}_j)-p_t(x)\Big]\\
    &= \sum_{i \in \Tc_t}\alpha_i\int K_{\sigma_i}(x-y)p_i(y)dy -p_t(x)\\
    &=\sum_{i \in \Tc_t}\alpha_i(K_{\sigma_i}*p_i)(x)-p_t(x),
\end{aligned}
\end{equation}
where $*$ denotes the convolution, and $p_i(\cdot)$ is the true density of batch $i$. 

Using $V(\cdot)$ to denote the variance operator, the estimator variance can be calculated as
\begin{equation}
\begin{aligned}
V(\hat{h}_t(x)) = \sum_{i\in \Tc_t} \alpha^2_iV(\hat{p}_i(x~; \sigma_i)),
    \end{aligned}
\end{equation}
due to the independence of batches, where
\begin{equation}\label{variance}
    V(\hat{p}_i(x~; \sigma_i)) = \frac{1}{n_i}\Big((K^2_{\sigma_i}*p_i)(x)-(K_{\sigma_i}*p_i)^2(x)\Big).
\end{equation}


The decomposition of the MSE of the "sliding window" estimator $\hat{h}_t$ is as follows
\begin{equation}\label{eq:MSEdecomp}
    \begin{aligned}
    MSE(\hat{h}_t,p_t) &= \E[(\hat{h}_t(x)-p_t(x))^2]\\
    &= V(\hat{h}_t(x)) + B^2(\hat{h}_t(x)).
    \end{aligned}
\end{equation}
Integrating above over $x$, we have
\begin{equation}
    MISE(\hat{h}_t,p_t) = \int MSE(\hat{h}_t,p_t) dx.
\end{equation}
Given the expressions of bias \eqref{bias} and variance \eqref{variance}, to calculate AMISE, we need to derive the Taylor approximations of the following quantities
\begin{equation}
    \begin{aligned}
    &(K^2_{\sigma_i}*p_i)(x)\\
    &(K_{\sigma_i}*p_i)(x).
    \end{aligned}
\end{equation}
First, we have
\begin{equation}
    \begin{aligned}
    (K^2_{\sigma_i}*p_i)(x) &= \int K^2_{\sigma_i}(x-y)p_i(y)dy\\
    &= \frac{1}{\sigma_i}\int K^2(z)p_i(x-\sigma_i z)dz\\
    &= \frac{p_i(x)}{\sigma_i}R(K) + o(1),
    \end{aligned}
\end{equation}
where we note that $p_i(x-\sigma_i z) = p_i(x)+o(1)$ holds, because $\sigma_i \rightarrow 0$ as $n_i \rightarrow \infty$. We also have that
\begin{equation}
    \begin{aligned}
    (K_{\sigma_i}*p_i)(x) &= \int K_{\sigma_i}(x-y)p_i(y)dy\\
    &= \int K(z)p_i(x-\sigma_i z)dz\\
    &= \int K(z)(p_i(x)-\sigma_i z p'_i(x) \\
    &+ \frac{1}{2}\sigma_i^2 z^2p''_i(x) + o(\sigma_i^2))dz\\
    &= p_i(x) + \frac{1}{2}\sigma_i^2 p''_i(x)\mu_2(K)+o(\sigma_i^2).
    \end{aligned}
\end{equation}
where we used the assumptions that $\int K(z)dz = 1$ and $\int zK(z) dz = 0$. Given the above asymptotic characterization of the quantities, we can rewrite the bias term \eqref{bias} as
\begin{equation}\label{eq:bias}
    \begin{aligned}
    B(\hat{h}_t(x)) = \sum_{i \in \Tc_t}\Big( \alpha_i b^{(t)}_i(x) +  \frac{1}{2}\alpha_i\sigma_i^2p''_i(x)\mu_2(K) +  o(\sigma^2_i)\Big).
    \end{aligned}
\end{equation}
We can also write the variance \eqref{variance} as
\begin{equation}\label{eq:var}
    V(\hat{h}_t(x)) = \sum_{i \in \Tc_t} \Big( \frac{\alpha_i^2}{n_i\sigma_i}R(K)p_i(x) + o(\frac{1}{n_i\sigma_i})\Big).
\end{equation}
We can now simplify the MSE \eqref{eq:MSEdecomp} as
\begin{equation}\label{eq:MSE}
    \begin{aligned}
    &MSE(\hat{h}_t,p_t)=\sum_{i \in \Tc_t} \Big( \frac{\alpha_i^2}{n_i\sigma_i}R(K)p_i(x) + o(\frac{1}{n_i\sigma_i})\Big)\\
    &+ \bigg(\sum_{i \in \Tc_t}\alpha_i b^{(t)}_i(x) + \sum_{i \in \Tc_t} \frac{1}{2}\sigma_i^2\alpha_ip''_i(x)\mu_2(K) + \sum_{i \in \Tc_t} o(\sigma^2_i)\bigg)^2.
    \end{aligned}
\end{equation}
Disregarding the terms that converge to zero and taking integral over $x$, we can derive an upper bound for AMISE as
\begin{equation}\label{eq:upperproof}
    \begin{aligned}
    AMISE(\hat{h}_t,p_t) &\leq \sum_{i \in \Tc_t}\frac{\alpha_i^2}{n_i\sigma_i}R(K)\\
    &+ (2|\Tc_t|-1)\sum_{i \in \Tc_t}\alpha^2_iR(b^{(t)}_i) \\
    &+ \frac{2|\Tc_t|-1}{4}\mu_2^2(K)\sum_{i \in \Tc_t}\alpha_i^2\sigma_i^4R(p_i''),
    \end{aligned}
\end{equation}
where the last two lines follow from the Cauchy-Schwarz inequality for the $2|\Tc_t|-1$ terms in the square. Note that $b_t^{(t)} = 0$ by definition. 
Observing that $|\Tc_t|=T_t$ completes the proof of Theorem \ref{the:upperbound}.
\end{proof}

Let us call the three lines in the right hand side of \eqref{eq:amise} as term $1$, term $2$, and term $3$, respectively. Term $1$ is due to the variance of the estimator, and terms $2$ and $3$ are the bias terms. Terms $1$ and $3$ are asymptotically vanishing in the sense that when $n_i\rightarrow\infty$, they both go to zero per condition \eqref{eq:band}. We can make several observations about the upper bound expression \eqref{eq:amise}. First, the dynamic density estimation with "sliding window" kernel density estimators will have a non-vanishing error term $2$, induced by keeping densities of various time stamps in the memory. {\rd We will later see in Corollary \ref{cor:consistent} that under optimal weight design, this term can also go to zero when $n_i\rightarrow\infty$.} Second, when the distribution evolution is mild (i.e., $R(b_i^{(t)})$ is small), there can be a theoretical advantage in including previous samples in the memory to reduce the variance term $1$. Later simulations will show this advantage can be significant in practice. Third, when the previous distributions are very different from the current distribution, it is desirable to only keep one batch (the current batch) in the memory, i.e., $\Tc_t = \{t\}$ and $T_t=1$. In this case, $R(b_t^{(t)})=0$ by definition \eqref{eq:diff} and the upper bound \eqref{eq:amise} exactly recovers the AMISE for the traditional kernel density estimator in \eqref{eq:mise}.
\subsection{Window Generator}

In the existing literature, kernel density estimators are modified using arbitrary "sliding windows" to adapt to the dynamic estimation. This approach performs better than the traditional kernel density estimator, as a static kernel density estimator works poorly for dynamic density estimation \cite{qian2019fast}. However, this heuristic approach lacks a theoretical justification. In fact, based on the theoretical upper bound of AMISE \eqref{eq:amise}, it is intuitive that the window size should depend on the density evolution to keep the AMISE small. For example, when the true density changes drastically, it is ideal to decrease the window size to adapt to the fast density change. Therefore, we propose a histogram-based window size generator that will allow the kernel density estimator to be adaptive to dynamic changes. 

We observe in \eqref{eq:amise} that compared to the static AMISE, the worst-case AMISE for dynamic density estimation depends on one more quantity, namely the difference function $b^{(t)}_i$. In principle, we can use this quantity as an indicator to adapt the dynamic kernel density estimator to the changes in the underlying density function.

We define a cutoff threshold to determine the number of batches (sliding window size) to be kept in the memory of the dynamic kernel density estimator. In doing so, we first define the temporal adaptive (TA) distance between two density functions. Here, we use histograms to approximate the density functions as true density functions are unavailable. We denote the number of bins in the histograms by $m$, set using the Sturges' rule \cite{sturges1926choice} 
\begin{equation}
    m = 1 + 3.322\log{n},
\end{equation}
where $n$ is the smallest batch size among all batches in the current memory. Sturges' rule is a widely adopted, simple binning algorithm in the literature. {\rd It is derived for normally distributed data. The user can choose other binning rules, such as Doane's rule \cite{doane1976aesthetic}, Scott's rule \cite{scott1979optimal}, or Freedman and Diaconis's rule \cite{freedman1981histogram} as appropriate. However, we note that all existing binning guidelines provide bins similar to Sturges' rule under low data volume (less than $200$) \cite{scott2015multivariate}.}

The temporal adaptive distance between two histograms $hist_i$ and $hist_j$ is  expressed as
\begin{equation}\label{eq:TAdistance}
    \|hist_i,hist_j\|_{TA} \triangleq \|\yb_i - \yb_j\|_2^2,
\end{equation} where $\|\cdot\|_2$ denotes the $\ell_2$ norm and $\yb_i$ is the probability mass vector on bins in batch $i$, i.e., $\|\yb_i\|_1 = 1$. 
This TA distance serves as a measure proportional to $\hat{R}(b_i^{(t)})$, the approximation of $R(b_i^{(t)})$ in \eqref{eq:amise}, i.e., 
\begin{equation}\label{eq:R}
    \hat{R}(b_i^{(t)}) \propto \|hist_i,hist_t\|_{TA}.
\end{equation}
To control the bias, one can set a cutoff threshold $s$ for the TA distance.  Upon receiving batch $t$, the number of batches to be kept in the memory can be set as $T_t$ that satisfies the following two inequalities
\begin{equation}\label{cut}
    \sum_{j = t-T_t}^{t-1} \|hist_j,hist_t\|_{TA} > s, \sum_{j = t-T_t+1}^{t-1} \|hist_j,hist_t\|_{TA} \leq s.
\end{equation}
Note that from a practical standpoint, the cutoff threshold $s$ should not be the only criterion for window selection, because when the true density goes through a long static period, it is possible that \eqref{cut} will induce a large memory window that exceeds the computational limit for real-time density estimation. Therefore, there should exist a hard cap $w$ to account for computational limits. Combining both considerations, the actual number of batches in the memory should be set as $\min(T_t,w)$.

\begin{remark}\label{re:1}
{\rd Note that the main purpose of cutoff value $s$ is to reduce the window size (and computation cost) when dealing with rapidly changing densities. The bias-variance decomposition suggests that including more batches in TAKDE can induce a lower variance (first term in equation \eqref{eq:var}) at the cost of increasing the bias (first term in equation \eqref{eq:bias}). 
Moreover, we will show in Corollary \ref{cor:consistent} that TAKDE is consistent regardless of window size $T_t$. Later,} synthetic data simulation {\rd also} suggests the empirical performance difference is not too sensitive to the cutoff value, so one can heuristically choose it in favor of fast processing rather than through intensive cross-validation.
\end{remark}

\begin{figure*}[t!]
\centering
\includegraphics[width = \textwidth,height = 0.4\textheight]{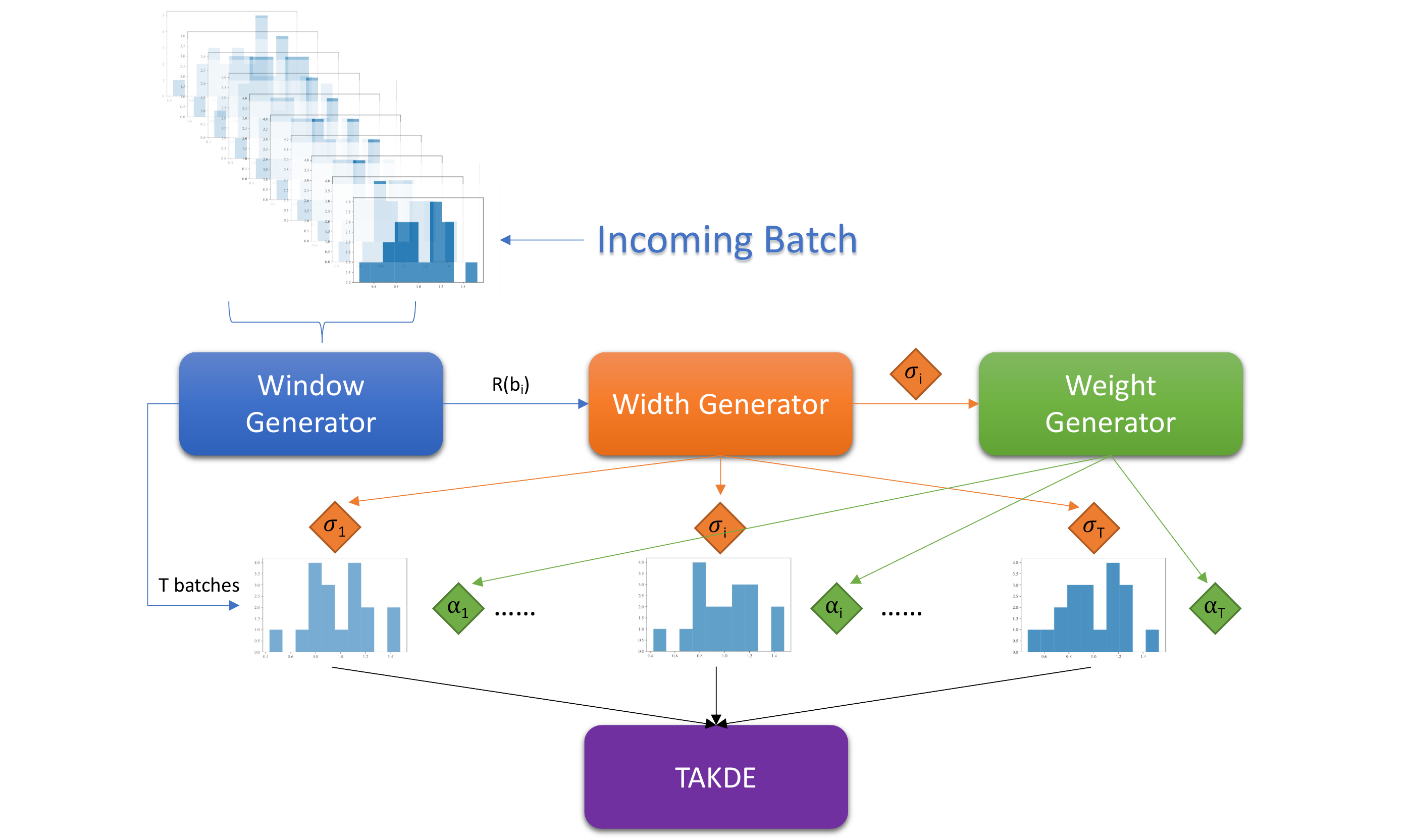}
\caption{TAKDE framework.}
\label{fig:flowchart}
\end{figure*}

\subsection{Bandwidth and Weight Generator}
The dynamic nature of the underlying true density makes it practically impossible to understand the actual difference functions and the second derivative of the true densities. However, using the AMISE upper bound in Theorem \ref{the:upperbound}, we can find theoretically optimal sequences for kernel bandwidths and weights, which in turn helps in the algorithm design. In view of Theorem \ref{the:upperbound}, we present the following corollary.

\begin{corollary}\label{cor:sequence}
The optimal sequences of weights and bandwidths that minimize the AMISE upper bound of the dynamic kernel density estimator are as follows
\begin{equation}
    \begin{aligned}
    \sigma_i &= \bigg[\frac{R(K)}{n_i\mu_2^2(K)R(p_i'')(2T_t-1)}\bigg]^{\frac{1}{5}},\\
    \alpha_i &= \frac{1/S_i}{\sum_{j \in \Tc_t}1/S_j},
    \end{aligned}
\end{equation}
where the sequence $S_i$ (with superscript $(t)$ omitted) is such that
\begin{equation}
    S_i = \frac{5R(K)}{4n_i\sigma_i}+(2T_t-1)R(b^{(t)}_i).
\end{equation}
\end{corollary}

\begin{proof}
Equation \eqref{eq:upperproof} shows that the upper bound on AMISE depends on the weight sequence $\alpha_i$ and the bandwidth sequence $\sigma_i$. Therefore, we can minimize the upper bound with respect to both of these parameters.

Differentiating with respect to $\sigma_i$ yields the following (optimal) sequence
\begin{equation}\label{bandwidth_seq}
    \sigma_i = \bigg[\frac{R(K)}{n_i\mu_2^2(K)R(p_i'')(2T_t-1)}\bigg]^{\frac{1}{5}}.
\end{equation}
We can find the optimal sequence of weights by simply solving the minimization of Lagrangian of \eqref{eq:upperproof} with the constraint $\sum \alpha_i = 1$ and incorporating \eqref{bandwidth_seq}. This will result in the following expression for the sequence $\alpha_i$
\begin{equation}
\begin{aligned}
    S_i &= \frac{5R(K)}{4n_i\sigma_i} + (2T_t-1)R(b^{(t)}_i)\\
    \alpha_i &= \frac{1/S_i}{\sum_{j \in \Tc_t}1/S_j},
\end{aligned}
\end{equation}
which completes the proof of
Corollary \ref{cor:sequence}.
\end{proof}

\begin{remark}
Corollary \ref{cor:sequence} provides some insights concerning the bandwidth and weight choices.
\begin{enumerate}
    \item The bandwidth sequence suggests that we should make the kernel more flexible as more batches of data points are included in the estimation. This aligns with the intuition from the traditional kernel density estimator, where the estimator can be more flexible with more sample points.
    \item The weight sequence provides the following insights. First, when the number of data points at a particular batch is considerably large, we should assign more weight to that batch with the hope of extracting more information to infer the current density. Second, the $R(b_i^{(t)})$ quantity provides a countermeasure to prevent us from assigning a large weight to data points coming from a very different distribution compared to the current batch. Third, we should assign more weights to the batches with larger kernel bandwidths, which means we are favoring smoother estimators in principle.
\end{enumerate}
\end{remark}

{\rd
\begin{corollary}\label{cor:consistent}
Under Assumptions \ref{assump:kernel}-\ref{assump:density}, the optimal weight sequence and kernel bandwidth sequence in Corollary \ref{cor:sequence} will ensure that for any $\epsilon > 0$,
\begin{equation}
    Pr(| \hat{h}_t-p_t |^2 > \epsilon) \rightarrow 0,
\end{equation}
as $n_i \rightarrow \infty$.
\end{corollary}

\begin{proof}
First, notice that following Corollary \ref{cor:sequence}, we have $\sigma_i \rightarrow 0$ and $\alpha_i \rightarrow 0$ for every batch except the last batch where $\alpha_t \rightarrow 1$ (since $R(b_t^{(t)}) = 0$) as $n_i \rightarrow \infty$. It is easy to verify that $\E[| \hat{h}_t-p_t |^2] \rightarrow 0$ under this bandwidth and weight sequence, based on the expression of the mean squared error in \eqref{eq:MSE}. Then, by Markov inequality, we have
\begin{equation}
    Pr(| \hat{h}_t-p_t |^2 > \epsilon) \leq \frac{\E[| \hat{h}_t-p_t |^2]}{\epsilon} \rightarrow 0.
\end{equation}
The proof is complete.
\end{proof}

Corollary \ref{cor:consistent} shows that TAKDE is weakly consistent as $n_i \rightarrow \infty$ regardless of $T_t$. This is rather intuitive as TAKDE can precisely recover the traditional KDE in this extreme case. 
}

\subsection{Kernel Bandwidth Selector}\label{sec:bandwidth}
The bandwidth sequence in Corollary \ref{cor:sequence} presents a principle for choosing the kernel bandwidth. However, the quantity $R(p_i'')$ is unknown in practice, and we still need to find a kernel bandwidth selector to calculate the actual kernel bandwidth values. There exist extensive studies for the choice of bandwidth in traditional kernel density estimation. One popular choice is the cross-validation approach \cite{bowman1984alternative,hall1991optimal,robert1976choice,rudemo1982empirical}. However, the computational  cost of cross-validation prohibits its application in high-frequency density estimation as every new batch of data points needs to be cross-validated for a new kernel bandwidth. 

Minimizing AMISE in \eqref{eq:mise} reveals a simple expression for the optimal kernel bandwidth. \cite{wand1994kernel} characterized the optimal kernel bandwidth based on \eqref{eq:mise} as follows
\begin{equation}\label{optimalsigma}
    \sigma_{AMISE} = \left[\frac{R(K)}{n\mu^2_2(K)R(p'')}\right]^{\frac{1}{5}}.
\end{equation}
We notice that \eqref{optimalsigma} coincides with the optimal kernel bandwidth sequence we derived in Corollary \ref{cor:sequence} except for a factor of $(2T_t-1)^{1/5}$. This relationship allows us to directly adopt existing kernel bandwidth selection methods for optimal AMISE.

Expression \eqref{optimalsigma} is still dependent on the unknown $R(p'')$, but there exist a number of studies that explore different methods for estimating $R(p'')$. For example, \cite{shimazaki2010kernel} approximates the AMISE objective function assuming the density is Poisson and then proceeds to find the minimizer as the optimal kernel bandwidth. However, this method is not applicable in real-time dynamic density estimation as the optimization process is expensive. \cite{qahtan2016kde} provides an iterative update framework by estimating $R(p'')$ through $R(\hat{p}'')$, which is the numerical square integration of the second derivative of the density estimator. This approach does not impose any strict assumption on the underlying distribution, which offers a robust estimation of $R(p'')$. However, the iterative algorithm still requires numerical operations like numerical derivatives and numerical integration, which may not be efficient enough for real-time density estimation.

In TAKDE, we adopt the normal rule introduced in \cite{silverman2018density}. Assuming the true density is Gaussian, the optimal kernel bandwidth can be approximated as follows
\begin{equation}\label{sigma_AMISE}
    \sigma_{AMISE} \approx c\hat{\sigma}n^{-\frac{1}{5}},
\end{equation}
where $c$ is the smoothness parameter depending on the kernel function and the underlying true density, and $\hat{\sigma}$ is the sample standard deviation of the data points. The normal rule is particularly appealing for the design of TAKDE due to its simple structure, which allows a direct plug-in of smoothness parameter $c$ and enables fast real-time processing.

There are two commonly used recommendations for the smoothness parameter $c$ in \eqref{sigma_AMISE}. The first choice given in \cite{wand1994kernel} is as follows
\begin{equation}\label{eq:normal}
    \sigma_{AMISE} \approx \left[\frac{8\pi^{1/2}R(K)}{3\mu_2^2(K)n}\right]^{\frac{1}{5}}\hat{\sigma},
\end{equation}
where $\hat{\sigma}$ is the estimated standard deviation assuming the true density is normal. The smoothness parameter $c$ of Gaussian Kernel in this setting is $(32/3)^{1/5}$.

\begin{algorithm}[ht]
	\caption{Temporal Adaptive Kernel Density Estimator (TAKDE)}\label{algo:TAKDE}
	{\bf Input:} 
	Kernel function $K(\cdot)$, cutoff value $s$, hard cap $w$, smoothness parameter $c$.
	
	{\bf For $t=1,2,\ldots$}
	
	\begin{algorithmic}[1]
	\STATE Receive new batch of data $\xb^{(t)}$ at time $t$.
	\STATE {\bf Window Generator:}
	Generate and record $hist_t$ and forget $hist_{t-w}$. Set $Distance = 0,T_t=0, \Tc_t=\emptyset$.
	
	{\bf While $T_t<w$:}
	\begin{equation}
	    Distance = Distance + \|hist_t,hist_{t-T_t}\|_{TA}
	\end{equation}
	{\bf Break If:}
	\begin{equation}
	    Distance > s,
	\end{equation}
	{\bf Else:}
	\begin{equation}
	    \Tc_t=\Tc_t\cup \xb^{(t-T_t)} ~~~~~~~T_t=T_t+1.
	\end{equation}
	{\bf Return:} $\Tc_t$ and $T_t$ and the sequence $\{\hat{R}(b_j^{(t)})\}_{j \in \Tc_t}$  where
	\begin{equation}
	    \hat{R}(b_j^{(t)}) = m\|hist_{j},hist_{t}\|_{TA}.
	\end{equation}
	
	\STATE {\bf Bandwidth Generator:} Receive the batch set $\Tc_t$.
	
	{\bf For $j \in \{t-T_t+1,\ldots,t\}$:}
	\begin{equation}
	    \sigma_j = \frac{c\hat{\sigma}_j}{((2T_t-1)n_j)^{1/5}},
	\end{equation}
	where $c$ is defined by the kernel bandwidth selector, $n_j=|\xb^{(j)}|$, and $\hat{\sigma}_j$ is the sample standard deviation of data in batch $j$.
	
	{\bf Return:} Bandwidth sequence $\sigma_j$.
	
	\STATE {\bf Weight Generator:}
	Receive bandwidth sequence $\sigma_j$ and the approximated $\hat{R}(b_j^{(t)})$ sequence. Let
	
	\begin{equation}\label{eq:alpha}
	    \begin{aligned}
	        \alpha_j &= \frac{1/S_j}{\sum_{i\in \Tc_t}1/S_i},\\
	        S_j &= \frac{5R(K)}{4n_j\sigma_j}+(2T_t-1)\hat{R}(b_j^{(t)}).
	    \end{aligned}
	\end{equation}
	
	{\bf Return:} Weight sequence $\alpha_j$.
	
	\end{algorithmic}
	
	{\bf Output:} 
	The Temporal Adaptive Kernel Density Estimator given as
	\begin{equation}
	    \begin{aligned}
	        \hat{h}_t(x) &= \sum_{j \in \Tc_t} \alpha_j\hat{p}_j(x~;\sigma_j),\\	   \hat{p}_j(x~;\sigma_j) &= \frac{1}{n_j}\sum_{i=1}^{n_j}K_{\sigma_j}(x-x^{(j)}_i).
	    \end{aligned}
	\end{equation}
\end{algorithm}

The second recommendation \cite{terrell1990maximal} comes from the upper bound of the AMISE-optimal kernel bandwidth using $beta(4,4)$ or triweight density function, that is,
\begin{equation}\label{eq:oversmooth}
    \sigma_{AMISE} \leq \left[\frac{243R(K)}{35\mu_2^2(K)n}\right]^{\frac{1}{5}}\hat{\sigma}.
\end{equation}
This bandwidth provides an oversmoothed density estimator that might not perform well with respect to metrics like log-likelihood or MSE. However, an oversmoothed density estimator is often preferred for real-world applications, because the results are visually plausible. In this case, the smoothness parameter $c$ of Gaussian Kernel is $(972/35\sqrt{\pi})^{1/5}$.

\begin{remark}
The only reason for adopting the normal rule in TAKDE is its computation simplicity. We must note that the weight sequence given in Corollary \ref{cor:sequence} is compatible with any existing  $R(p'')$ approximation method.
\end{remark}

\subsection{Algorithm Design}

In this subsection, we present the final form of TAKDE. The algorithm requires as input a cutoff value $s$, a hard cap $w$, a smoothness parameter $c$, and a kernel function $K$. Upon receiving the batch of data points at time $t$, the window generator decides the set of batches $\Tc_t$ to be used for the density estimation. The window generator will also return the sequence of approximated $\hat{R}(b_j^{(t)})$ as in \eqref{eq:R} for all batches in the memory. Then, all batches within the memory will be fed into the bandwidth generator to generate the sequence of kernel bandwidths $\sigma_j$ as in Corollary $\ref{cor:sequence}$. Then, the approximated $\hat{R}(b_j^{(t)})$ and bandwidth sequence $\sigma_j$ will be fed into the weight generator to generate the sequence $\alpha_j$ as in Corollary \ref{cor:sequence}. Finally, all parameters will be put together to generate a proper kernel density estimator for estimating the density at time $t$. Fig. \ref{fig:flowchart} illustrates the workflow of TAKDE. The algorithmic presentation of TAKDE is outlined in Algorithm \ref{algo:TAKDE}.

\section{Experiment}\label{sec:experiment}
We now present numerical experiments to verify the efficiency of TAKDE both on synthetic data and real-world data. All experimental results established in this section are based on Gaussian kernel function.

\subsection{Algorithm Design Evaluation}
Before we compare TAKDE with other established benchmark algorithms, we evaluate the design of TAKDE on synthetic data. The specific question that we aim to address is that whether our proposed weighting scheme, derived from the AMISE upper bound, outperforms other heuristic weight sequences such as uniform (or average) weighting and exponentially decaying weighting.


\begin{figure*}[t!]
\centering
\includegraphics[width = \textwidth, height=0.4\textheight]{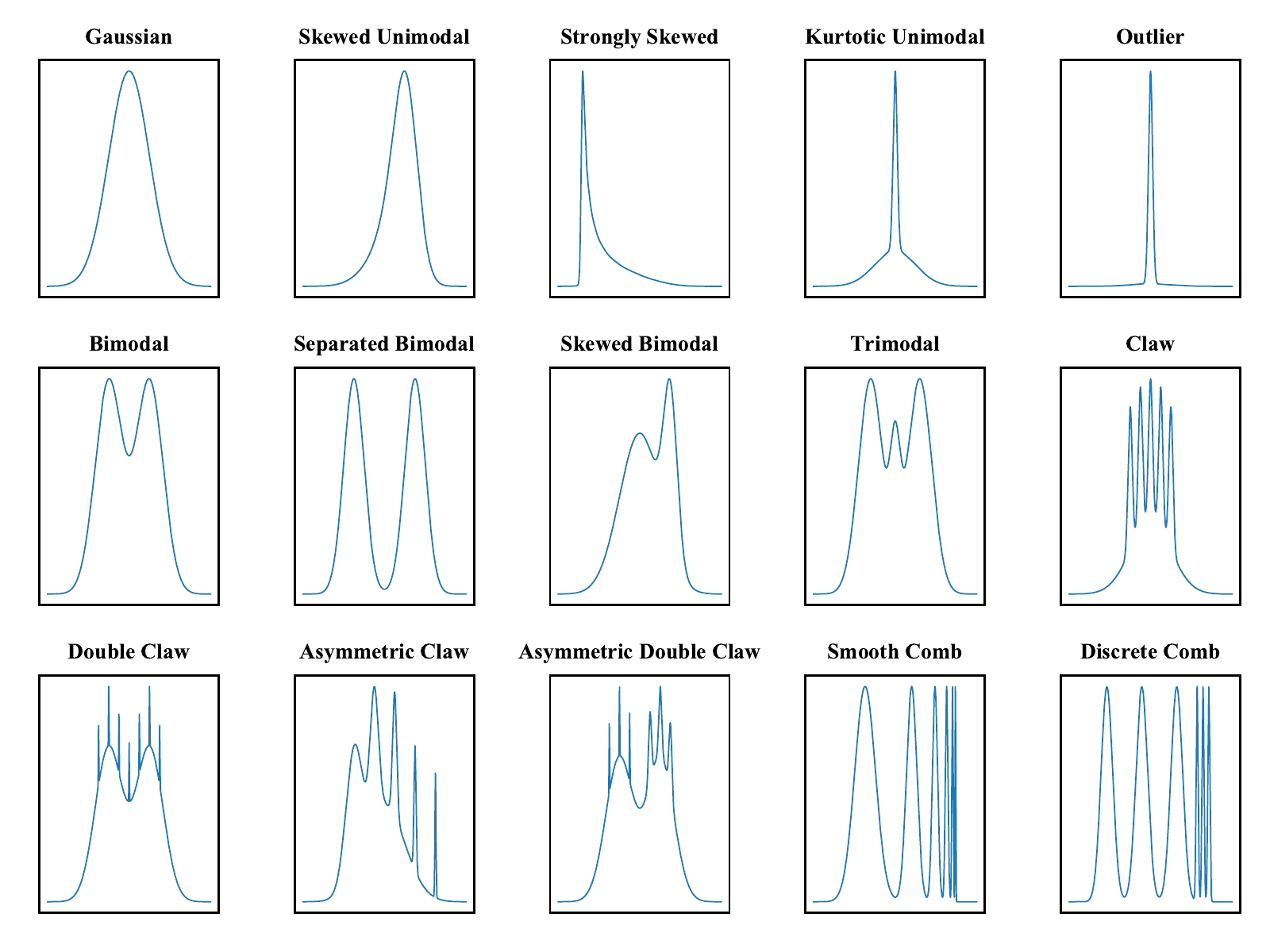}
\caption{The $15$ Gaussian mixture densities used in the synthetic dataset design.}
\label{fig:densities}
\end{figure*}

\subsubsection{Synthetic Dataset Design}
We create a synthetic dataset to test the performance of TAKDE in dynamic density estimation. We design the synthetic dataset following some general principles.
\begin{enumerate}
    \item The true densities involved in the generation of the dataset need to have analytical forms and have already been established in the literature.
    \item Each batch of data points has a size in the range of $[5, 20]$, so that the batches do not differ too drastically in terms of the data amount.
    \item The number of testing points for all batches should be the same for comparison purposes.
    \item The dynamics of the underlying densities varies for different batches.
\end{enumerate}
Following the above principles, we adopt the $15$ Gaussian mixture densities, recommended by \cite{marron1992exact}, as the baseline densities for our synthetic dataset design. The $15$ densities are shown in Fig. \ref{fig:densities}. 

To design the true density, we first consider $14$ sections, where each section consists of multiple batches.  
Let us denote the $15$ Gaussian mixtures with $g_1(x),\ldots,g_{15}(x)$ and represent the $14$ sections with $\Sc_1,\ldots,\Sc_{14}$, where $|\Sc_1|+\ldots+|\Sc_{14}|$ equals to the total number of batches in the dataset. To be specific, section $\Sc_i$ has $|\Sc_i|$ consecutive batches of data points in it, and the first batch of data in section $\Sc_{i+1}$ will start after the last batch in section $\Sc_i$. For batch $i$ in section $j$, where $1 \leq i \leq |\Sc_j|$, the density function $h_i^{(j)}(x)$ is defined as follows
\begin{equation}
    h_i^{(j)}(x) = \frac{|\Sc_j|-i+1}{|\Sc_j|} g_j(x) + \frac{i-1}{|\Sc_j|} g_{j+1}(x).
\end{equation}
To be consistent with our previous notation, $h_i^{(j)}(x)=p_{t_{ij}}(x)$ for $t_{ij}=|\Sc_1|+\ldots+|\Sc_{j-1}|+i$. Notice that in section $j$, the $j$-th Gaussian mixture linearly transforms to the $j+1$-th Gaussian mixture. After we move on to section $j+1$, none of previous Gaussian mixtures $g_1(x),\ldots,g_{j}(x)$ will appear in the section. 
Given the density of batch $i$ in section $j$, we sample a random number between $5$ to $20$ as the number of training points and $500$ for testing points to perform the comparison. To account for the randomness in partitioning the batches into $14$ sections and the randomness in samples, we generate $300$ synthetic datasets for Monte-Carlo simulations. 

\subsubsection{TAKDE Evaluation}
We now compare the weight generator in TAKDE with two heuristic approaches in the literature. One approach is to assign uniform weights to the batches, assuming older data points are of the same importance as the new data points, and the other one is to assign exponentially decaying weights, assuming the new points are much more important \cite{heinz2008cluster,qahtan2016kde}. 
To ensure a fair comparison, we only change the weight generator of TAKDE to uniform and exponential weighting, and we keep the other components of the algorithm unchanged. 
The uniform weight sequence is set as follows
\begin{equation}
    \alpha_j = \frac{1}{T_t}, \text{$\forall j\in \{t-T_t+1,\ldots,t\}$}. 
\end{equation}
The exponential weight sequence is set as follows
\begin{equation}
    \begin{aligned}
        \alpha_{j} &= (1-e)e^{t-j}, \text{$\forall j\in \{t-T_t+2,\ldots,t\}$},
    \end{aligned}
\end{equation}
and $\alpha_{t-T_t+1} = e^{T_t-1}$, where $e$ is the decay ratio. We compare the above to $\alpha_j$ corresponding to the expression in \eqref{eq:alpha}. In our simulation, $e=0.9$ in general yields the best result under different settings; therefore, the decay ratio for exponential weight sequence is set to $e=0.9$.

Our comparison is performed under several kernel bandwidth selectors, including the normal selector and oversmooth selector mentioned in Section \ref{sec:bandwidth} and under various cutoff values. 

\begin{figure*}[t!]
\centering
\includegraphics[width = \textwidth, height=0.3\textheight]{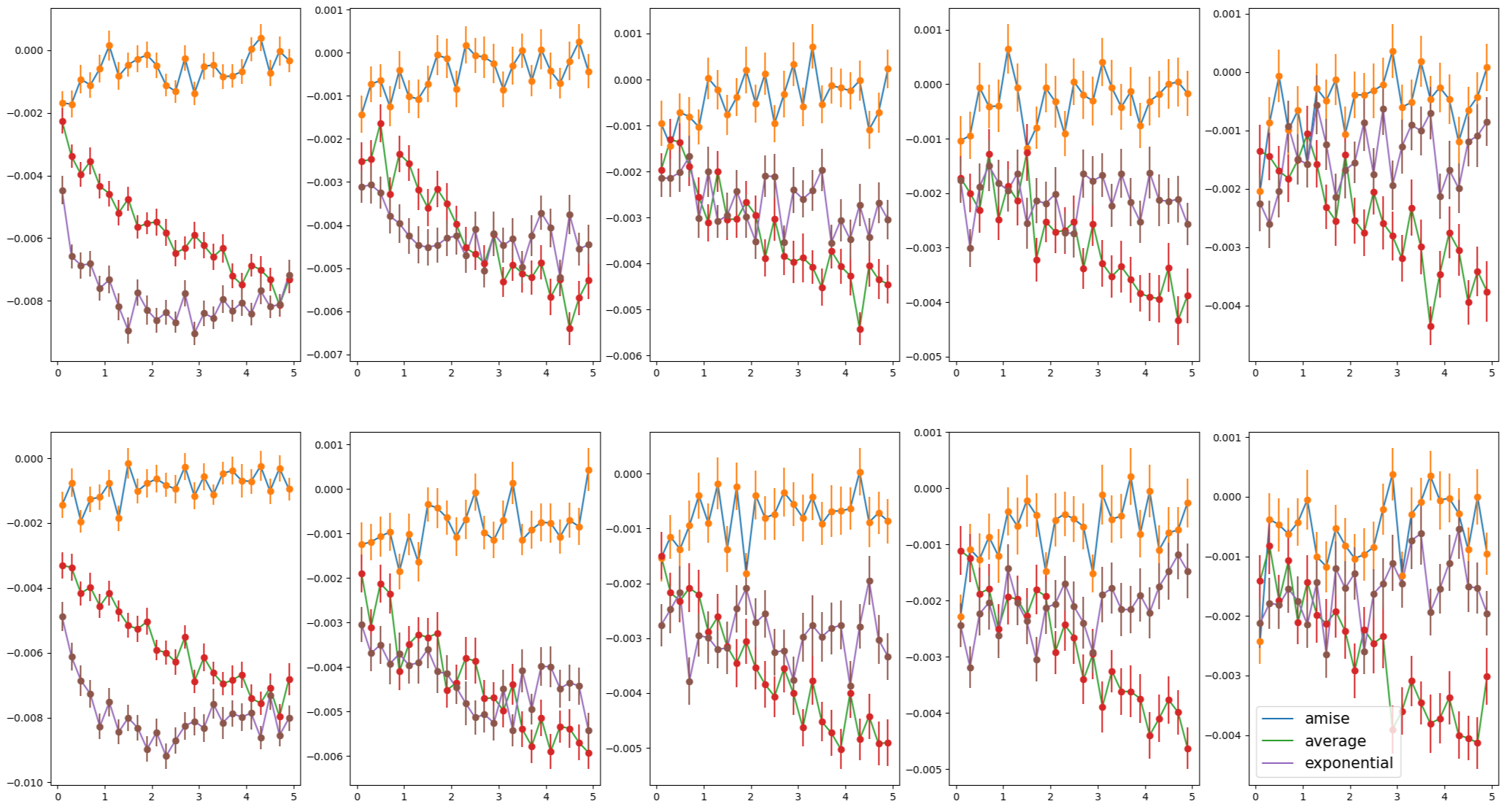}
\caption{The test log-likelihood comparison between TAKDE vs. the heuristic approaches. The x-axis represents the cutoff value and the y-axis represents the test log-likelihood. The first row shows the result under normal bandwidth selector and the second row shows the result under oversmooth bandwidth selector. In each row, the plots from left to right represent the simulation results using synthetic datasets with $100,200,300,400,500$ batches of data.}
\label{fig:synthetic_comparison}
\end{figure*}

\begin{figure*}[t!]
\centering
\includegraphics[width = \textwidth, height=0.26\textheight]{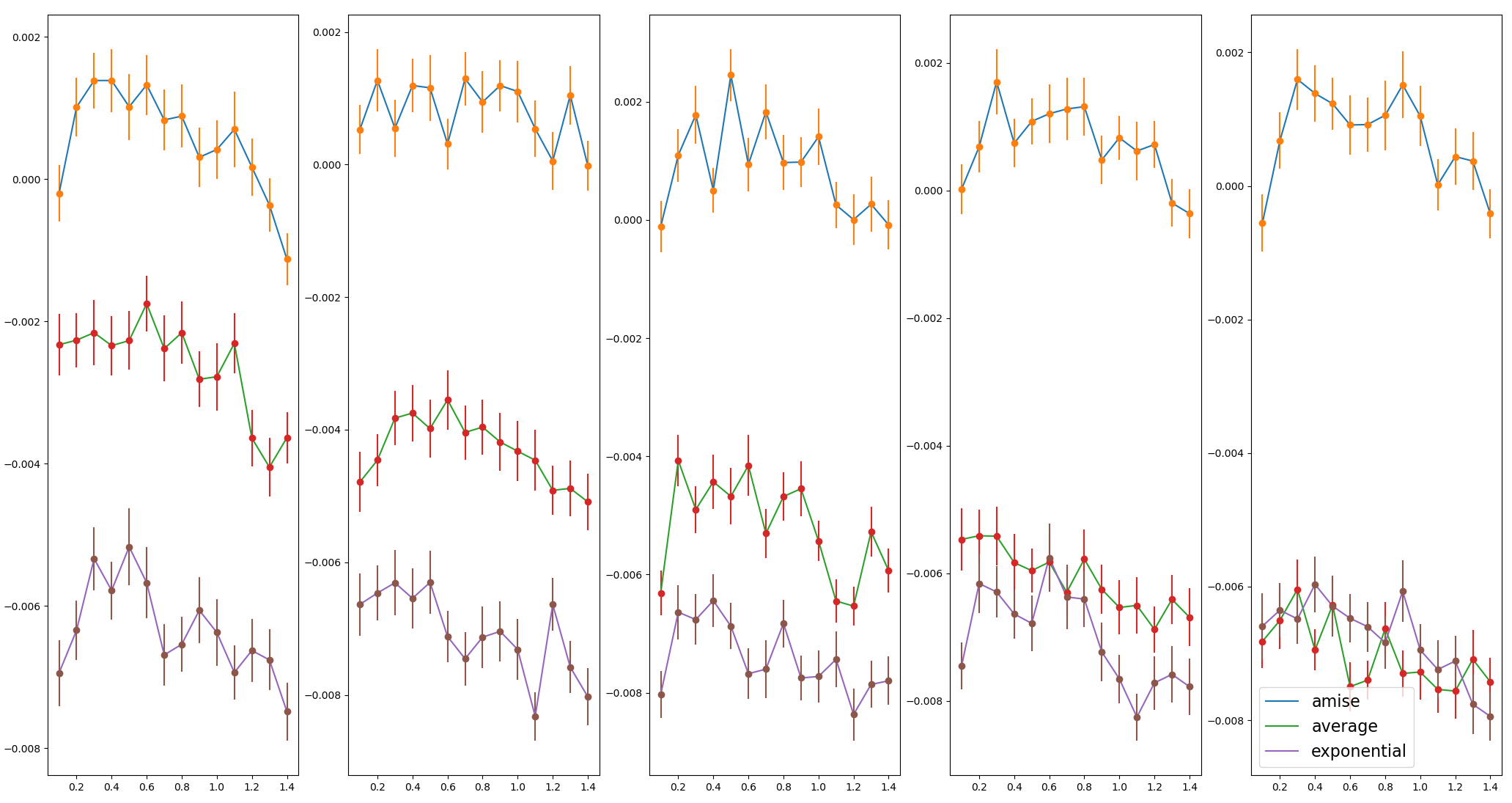}
\caption{The test log-likelihood comparison between TAKDE vs. the heuristic approaches over different bandwidth selectors. The x-axis represents the value of the smoothness parameter $c$. The y-axis represents the test log-likelihood. Each plot from left to right represents the simulation conducted with cutoff values from $1$ to $5$.}
\label{fig:synthetic_cv}
\end{figure*}

\begin{table*}
    \centering
    \caption{The best experimental settings for different benchmark algorithms in different datasets.}
    \label{tab:settings}
    \begin{tabular}{|c|c|c|c|c|}
        \hline
        Algorithm &  Noise Parameter $\alpha_1$ & Noise Parameter $\alpha_2$ & Smoothness Parameter $c$ & Cutoff Value (Window Size)\\
        \hline
        \multicolumn{5}{|c|}{TEM}\\
        \hline
        KDE  & - & - & $1.34$ & - \\
        B-spline & $0.66$ & $0.04$ & - & - \\
        KDEtrack & - & - & $0.45$ & $1(16)$ \\
        TAKDE & - & - & $0.15$ & $1(16)$ \\
        \hline
        \multicolumn{5}{|c|}{ECG}\\
        \hline
        KDE  & - & - & $0.98$ & - \\
        B-spline & $0.82$ & $0.05$ & - & - \\
        KDEtrack & - & - & $0.1$ & $1(60)$ \\
        TAKDE & - & - & $0.7$ & $1(60)$ \\
        \hline
        \multicolumn{5}{|c|}{Wafer}\\
        \hline
        KDE  & - & - & $0.22$ & - \\
        B-spline & $0.96$ & $0.06$ & - & - \\
        KDEtrack & - & - & $1.05$ & $1(20)$ \\
        TAKDE & - & - & $0.15$ & $1(20)$ \\
        \hline
        \multicolumn{5}{|c|}{Earth}\\
        \hline
        KDE  & - & - & $0.4$ & - \\
        B-spline & $0.81$ & $0.05$ & - & - \\
        KDEtrack & - & - & $0.8$ & $0.2(15)$ \\
        TAKDE & - & - & $0.05$ & $1.4(90)$ \\
        \hline
        \multicolumn{5}{|c|}{Star}\\
        \hline
        KDE  & - & - & $0.9$ & - \\
        B-spline & $0.30$ & $0.02$ & - & - \\
        KDEtrack & - & - & $1$ & $0.3(20)$ \\
        TAKDE & - & - & $0.35$ & $1.8(38)$ \\
        \hline
    \end{tabular}
\end{table*}

\begin{table*}
\begin{center}
\caption{Mean test log-likelihood on five real datasets.}\label{tab:bench}
\begin{tabular}{ |c|c|c|c|c|c| } 
 \hline
 Algorithm & TEM & ECG & Wafer & Earth & Star\\
 \hline
 KDE & $-0.026\pm0.0001$ & $0.060\pm0.00002$ & $0.0229\pm0.0007$ & $0.048\pm0.0002$ & $0.0078\pm0.00002$\\
 B-spline Kalman Filter & $0.171\pm0.0062$ & $1.580\pm0.0011$ & $1.204\pm0.0034$ & $1.324\pm0.0051$ &$0.685\pm0.0024$\\
 KDEtrack & $0.245\pm0.0057$ & $1.095\pm0.0009$ & $0.866\pm0.0018$ & $0.915\pm0.0012$ &$0.640\pm0.0007$\\
 TAKDE(normal) & $0.130\pm0.0016$ & $\mathbf{1.639\pm0.0004}$ & $\mathbf{1.530\pm0.0015}$ & $1.247\pm0.0017$ &$\mathbf{0.696\pm0.0007}$\\
 TAKDE(cor) & $\mathbf{0.246\pm0.0022}$ & $\mathbf{1.625\pm0.0010}$ & $\mathbf{1.627\pm0.0017}$ & $\mathbf{1.331\pm0.0026}$ & $\mathbf{0.705\pm0.0008}$\\
 TAKDE & $\mathbf{0.362\pm0.0036}$ & $\mathbf{1.648\pm0.0009}$ & $\mathbf{1.848\pm0.0025}$ & $\mathbf{1.504\pm0.0026}$ & $\mathbf{0.710\pm0.0012}$\\
 \hline
\end{tabular}
\end{center}
\end{table*}

\begin{table*}
\begin{center}
\caption{Run-time comparison (seconds) on five real datasets.}\label{tab:runtime}
\begin{tabular}{ |c|c|c|c|c|c| } 
 \hline
 Algorithm & TEM & ECG & Wafer & Earth & Star\\
 \hline
 B-spline Kalman Filter & $7.08$ & $4.099$ & $0.379$ & $0.907$ & $1.752$\\
 KDEtrack & $5.461$ & $4.712$ & $1.542$ & $1.569$ & $14.85$\\
 TAKDE & $\mathbf{0.378}$ & $\mathbf{0.557}$ & $\mathbf{0.114}$ & $\mathbf{0.704}$ & $\mathbf{0.851}$\\
 \hline
\end{tabular}
\end{center}
\end{table*}

First, we consider normal bandwidth selector \eqref{eq:normal} and oversmooth bandwidth selector \eqref{eq:oversmooth}. For each bandwidth selector, we conduct the comparison with datasets having from $100$ to $500$ batches of data to reflect different underlying dynamics. Notice that for the data with $100$ batches, the dynamic change is more drastic than that of the data with $500$ batches.

The simulation result is shown in Fig. \ref{fig:synthetic_comparison}. We can observe that TAKDE with AMISE-based weight sequence dominates the uniform and exponential weight sequences in terms of the test log-likelihood. We also see that when using the heuristic weight sequences, increasing the memory (i.e., larger cutoff value) mostly exacerbates the density estimation performance. The results show that the performance difference between TAKDE and other two methods is larger when the total number of batches is smaller. This suggests that TAKDE with AMISE-based weight sequence is better at adapting to more drastic dynamic changes. The smaller differences in $500$-batch simulations are consistent with our theoretical results, where the weighting sequence in Corollary \ref{cor:sequence} gets closer to uniform weighting as $R(b_i^{(t)})$ converges to $0$, equivalent to a static density estimation. We observe that changes in the cutoff value do not have a significant effect on TAKDE performance compared to others. This verifies our discussion in Remark \ref{re:1}.

Second, we conduct the comparison using a synthetic dataset with 100 batches of data for different bandwidth selectors, i.e., varying the smoothness parameter $c$ in \eqref{sigma_AMISE}. The simulation results are shown in Fig. \ref{fig:synthetic_cv}. Again, we observe the same performance trend for the algorithms. These simulations empirically verify that the performance advantage of our proposed weight sequence against the heuristic weight sequences is robust to different kernel bandwidths and different window sizes. 

\subsection{Comparison with Benchmark Algorithms}

Next, we compare TAKDE with three density estimation methods on real-world datasets. We consider both the mean test log-likelihood and the run-time to show the advantages of TAKDE.

\subsubsection{Benchmark Algorithms}

\begin{enumerate}
    \item {\bf Kernel Density Estimator (KDE)}: The first benchmark algorithm is the traditional kernel density estimator. The main reason to include kernel density estimator in the comparison is to show why a traditional density estimator is not ideal for dynamic density estimation. The kernel density estimator is formulated as \eqref{eq:kde}. The bandwidth selector is
    \begin{equation}
        \sigma = c\hat{\sigma}n^{-\frac{1}{5}},
    \end{equation}
    where we use cross-validation to choose $c$ (rather than the actual bandwidth) for easy comparison with TAKDE.
    \item {\bf B-spline Kalman Filter (BKF) \cite{qian2019fast}}: B-spline Kalman filter models the underlying density function as a count measure defined on the partitions of the density support. The density estimator is defined as
    \begin{equation}
        \hat{p}(x) = \frac{1}{C}\exp{\sum_{i=1}^m \beta_i B_i(x)},
    \end{equation}
    where $C$ is the normalization constant calculated with numerical integration, $m$ is the number of partitions, and $B_i(x)$ are the B-spline bases. The algorithm updates its states $\beta_i$ using a B-spline matrix evaluated on the centers of the density support partitions and the count vector at each batch.
    \item {\bf KDEtrack \cite{qahtan2016kde}}: KDEtrack partitions the support of the density using a collection of grid points. The set of grid points and the density values at the grid points are updated after each new batch of data points is received and evaluated. The density evaluation at a test point will be the linear interpolation at the test point using the closest grid points.
\end{enumerate}

\begin{remark}
We do not include the M-kernel and LRKDE methods since \cite{qahtan2016kde} has showed that KDEtrack is superior to these two methods.
\end{remark}

\subsubsection{Datasets}

\begin{itemize}
    \item {\bf In situ TEM video data}: The first dataset we use is in situ TEM dataset introduced in Section \ref{sec:introduction}. It is the $76.6$ second in situ TEM video published in \cite{zheng2009observation}. It has a total of $1149$ frames of images and $5-20$ particle counts in each frame.
    
    \item {\bf CinCECGTorso (ECG) data}: CinCECGTorso dataset is an ECG dataset taken from multiple torso surface sites of four patients from the Computers in Cardiology challenges. This dataset is available on UCR time-series data archive \cite{UCRArchive2018}.

    \begin{figure*}[t!]
\centering
\includegraphics[width = \textwidth, height=0.4\textheight]{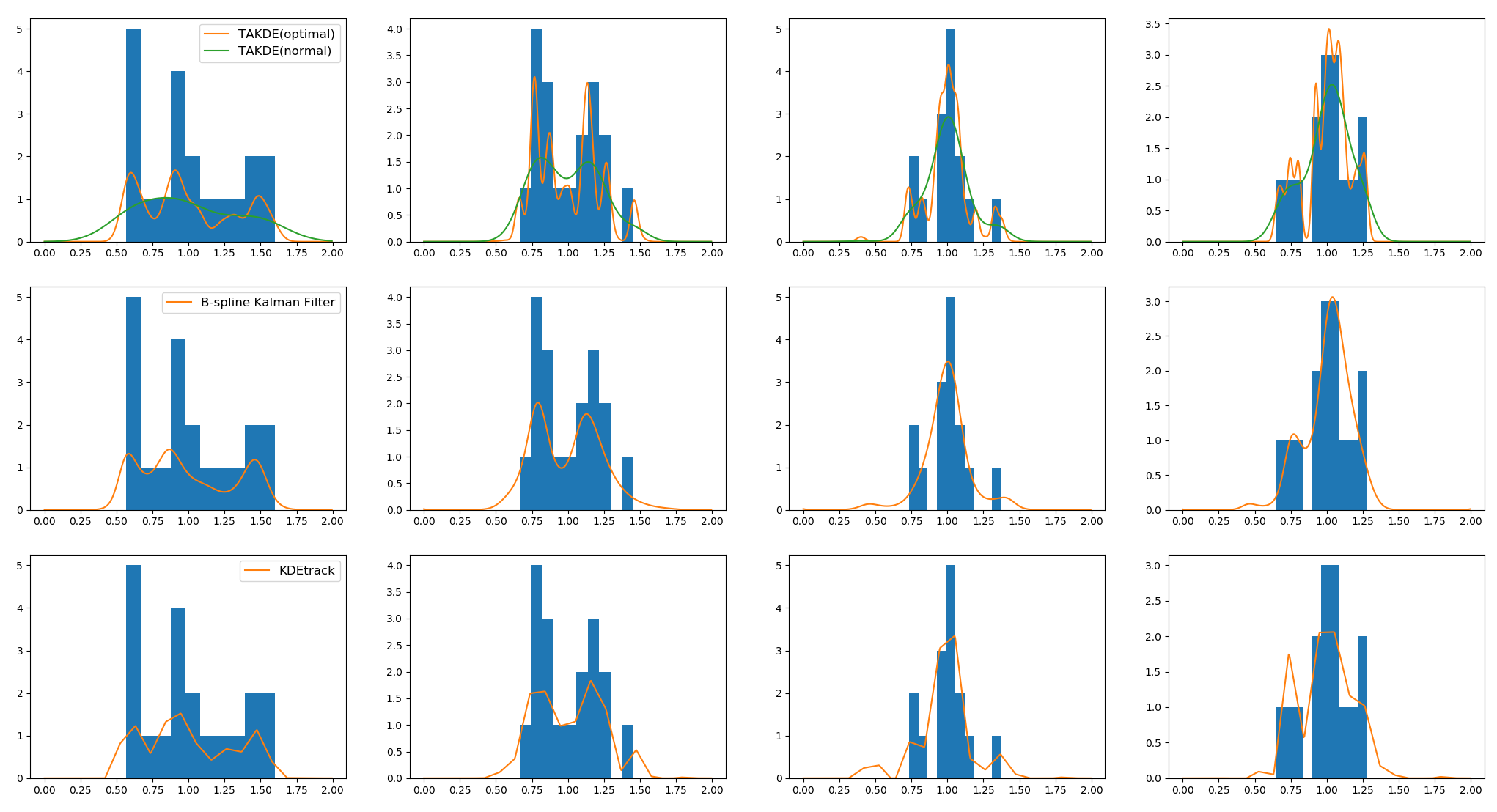}
\caption{Visualization of the density estimators on the TEM dataset. The first row shows TAKDE at its normal setting and optimal setting. The second row shows B-spline Kalman Filter at its optimal setting. The third row shows KDEtrack at its optimal setting. Figures from left to right represent the estimation at time stamps $225$, $450$, $675$, and $900$, respectively.}
\label{fig:visual}
\end{figure*}
    
 The dataset consists of ECG measurements of four patients. We use the ECG signal sequence of one person to highlight the density dynamics over time. Note that simulations on all four patients yield similar results. There are $342$ ECG signals (data points) available at each batch, and there are a total of $1639$ batches of data points over time. The batches are collected at $2$-kHz frequency, which requires the density estimator to be updated $2000$ times per second. For each batch of data points at a certain time stamp, we randomly sample $5$ to $20$ data points to train and use the rest of the data points to evaluate the algorithms. The number of training data points at each batch is determined only once throughout all the Monte-Carlo simulations. However, the set of training points are sampled randomly in each Monte-Carlo simulation.
    
    \item {\bf Wafer data}: Wafer dataset is a collection of sensor readings in a semiconductor wafer manufacturing process over time, available on UCR time-series data archive \cite{UCRArchive2018}. Unlike the previous two datasets, a wafer manufacturing process is a rather slow process that could span over $10$ weeks. However, this dataset is still illustrative for evaluating the accuracy of TAKDE. We use the readings in the normal state wafer manufacturing process to conduct our analysis. There are $600$ readings (data points) available at each batch, and there are a total of $152$ batches of data points over time. Again, we adopt the same train-test split approach as in the ECG dataset.

{\rd
    \item {\bf Earthquakes (Earth) data}: The earthquake dataset is a sensor reading dataset from Northern California Earthquake Data Center available on UCR time-series data archive \cite{UCRArchive2018}. It consists of $461$ readings at each batch with a total of $512$ batches.
    
    \item {\bf StarLight Curves (Star) data}: The starlight curves dataset consists of time-series sensor readings on the brightness of a collection of celestial objects. It is also available on UCR time-series data archive \cite{UCRArchive2018}. This dataset includes the readings of $1000$ celestial objects at each batch with a total of $1024$ batches.
    }
\end{itemize}

\subsubsection{Experimental Settings}

In comparing across different density estimators, we only present the best performance of B-spline Kalman filter, where the noise prior parameters are cross-validated using a grid search with an interval size of $0.01$. For the traditional kernel density estimator, we report its best performance, but even that is significantly inferior to other density estimators. For KDEtrack and TAKDE, we report the best settings performances (in terms of smoothness parameter $c$ and cutoff value $s$). Notice we do not adopt the iterative bandwidth update in KDEtrack for the computation reason explained in Section \ref{sec:bandwidth}, but instead we use the same bandwidth generator as in TAKDE. All the simulations are conducted over $100$ Monte-Carlo simulations for random training-testing splits to generate the standard errors of the performance. The performance metric is the mean test log-likelihood of the test points.

\subsubsection{Performance}

The parameter settings leading to respective best performance for all benchmark algorithms are shown in Table \ref{tab:settings}.  These settings are cross-validated using the first  $10\%$ batches of each dataset ($20\%$ for Wafer {\rd and Earth} dataset).

The results are tabulated in Table \ref{tab:bench}. TAKDE tagged with "(normal)" represents the performance achieved with smoothness parameter recommended in equation \eqref{eq:normal} (normal bandwidth selector) and the optimal cutoff in Table \ref{tab:settings}. TAKDE tagged with "(cor)" represents the performance achieved by TAKDE under KDEtrack best settings in terms of cutoff value and smoothness parameter. As we can observe, TAKDE dominates all other benchmark algorithms in terms of test log-likelihood by a large margin. TAKDE is also robust with respect to different cutoff values and different smoothness parameters, as it dominates all  other benchmark algorithms even under the best settings for KDEtrack. The only exception is TAKDE with normal bandwidth selector on the TEM dataset. The underlying reason is that the   low data volume available at different batches (training and testing combined) forces the "true" density distribution at each time stamp to an average of Dirac measures, which is far from the normal assumption of the normal bandwidth selector.

The run-time comparisons are shown in Table \ref{tab:runtime}. The values represent the time used for executing the density estimation for all test data points in all batches. We can observe that in addition to being more accurate than the benchmark algorithms, TAKDE is also much faster in speed as it requires negligible calculations in addition to kernel density evaluation. The computation advantage makes a huge difference for the ECG dataset in particular, as the other two benchmark algorithms do not run nearly fast enough to catch up with the $2$kHz data collection rate.

\subsection{Visual Examination}

In this subsection, we visualize the previously compared density estimators.
We pick the time stamps $\{225,450,675,900\}$ in $1150$ batches of data in the TEM dataset for visualization. The results are shown in Fig. \ref{fig:visual}. As we can observe, TAKDE at its optimal setting (for test log-likelihood) yields a more flexible model compared to other algorithms. TAKDE with normal smoothness parameter yields the smoothest model among all. Our results in Table \ref{tab:bench} also show that the normal smoothness parameter can achieve estimation performance close to the optimal setting while yielding smooth density functions that facilitate easy interpretation. For this reason, in most real-world applications that do not place estimation accuracy as their first priority, we do recommend using the normal smoothness parameter \eqref{eq:normal} to avoid cross-validation.

\section{Conclusion}\label{sec:conclusion}

In this paper, we established a theoretical AMISE upper bound expression for the "sliding window" kernel density estimator in dynamic density estimation. We proposed the temporal adaptive kernel density estimator that maintains the fast processing advantage of the "sliding window" kernel density estimator, while being theoretically optimal under the worst-case AMISE. We provided extensive numerical simulations to verify that TAKDE is superior to state-of-the-art real-time dynamic density estimators in terms of the mean test log-likelihood. TAKDE also dominated these algorithms in terms of achieving smaller run-times.

The proposed weight sequence is reminiscent of the attention mechanism in a transformer neural network for sequence re-weighting \cite{vaswani2017attention}. Considering the massive success of transformers in different fields, one of the future research directions is to see whether learning the weight sequence through the attention mechanism can result in a better performance.

Note that TAKDE in its current state only works for univariate density estimation. Thus, another future direction is to extend it to multivariate density cases.

\ifCLASSOPTIONcompsoc
  \section*{Acknowledgments}
\else
  \section*{Acknowledgment}
\fi

The authors gratefully acknowledge the support of NSF Award \#2038625 as part of the NSF/DHS/DOT/NIH/USDA-NIFA Cyber-Physical Systems Program.

\ifCLASSOPTIONcaptionsoff
  \newpage
\fi

\bibliographystyle{IEEEtran}
\bibliography {bibliography.bib}

\begin{IEEEbiography}
	[{\includegraphics[width=1in,height=1.25in,clip,keepaspectratio]{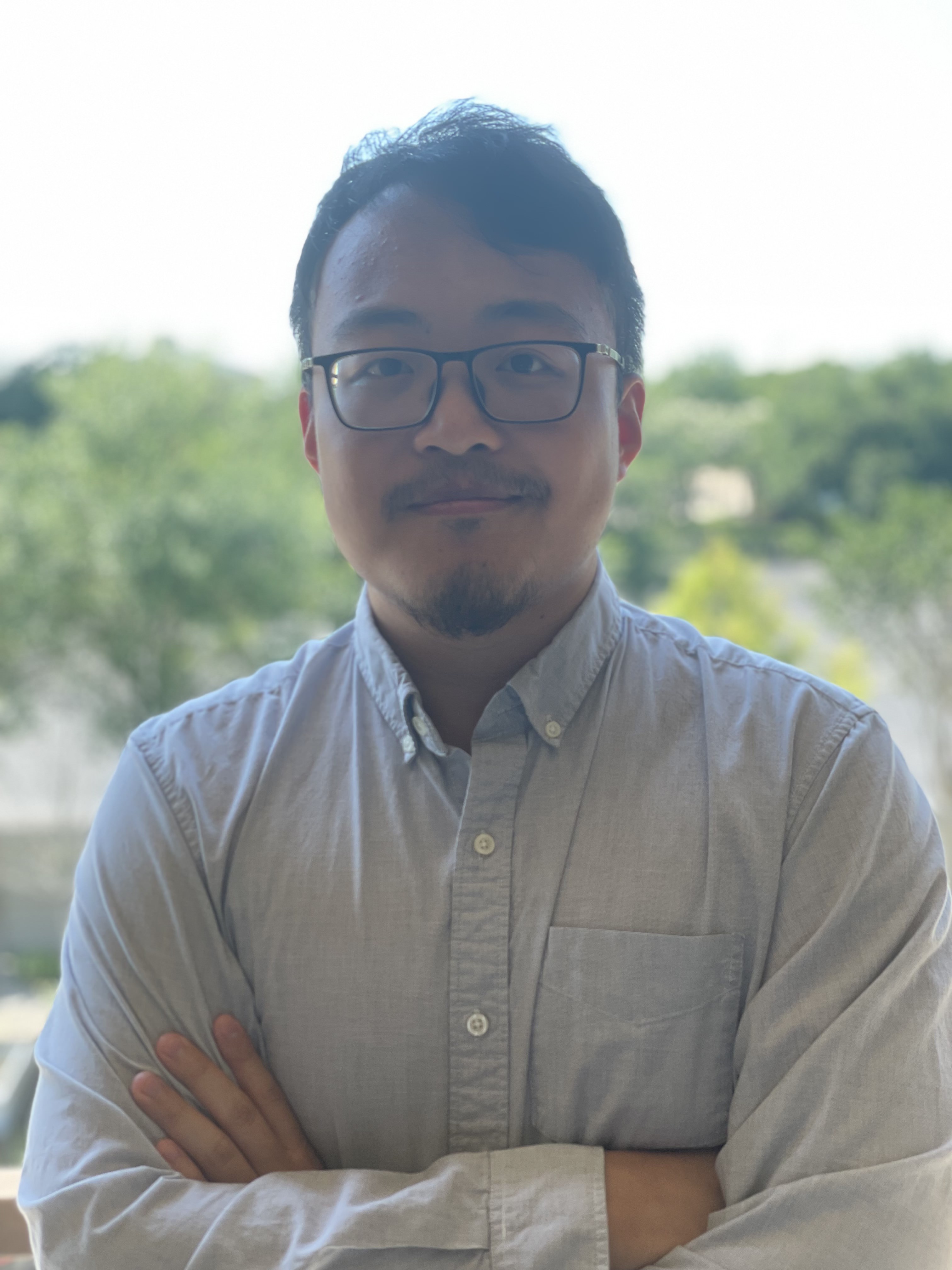}}] {Yinsong Wang} received his B.S. degree in Mechanical Engineering from Shandong University, China, in 2017, and his M.S. degree in Manufacturing System Engineering and Management from The Hong Kong Polytechnic University, Hong Kong, in 2018. He is currently working toward a Ph.D. degree in Industrial Engineering at Northeastern University. His research interests include machine learning, data science, and kernel methods.
	\end{IEEEbiography}

\begin{IEEEbiography}[{\includegraphics[width=1in,height=1.25in,clip,keepaspectratio]{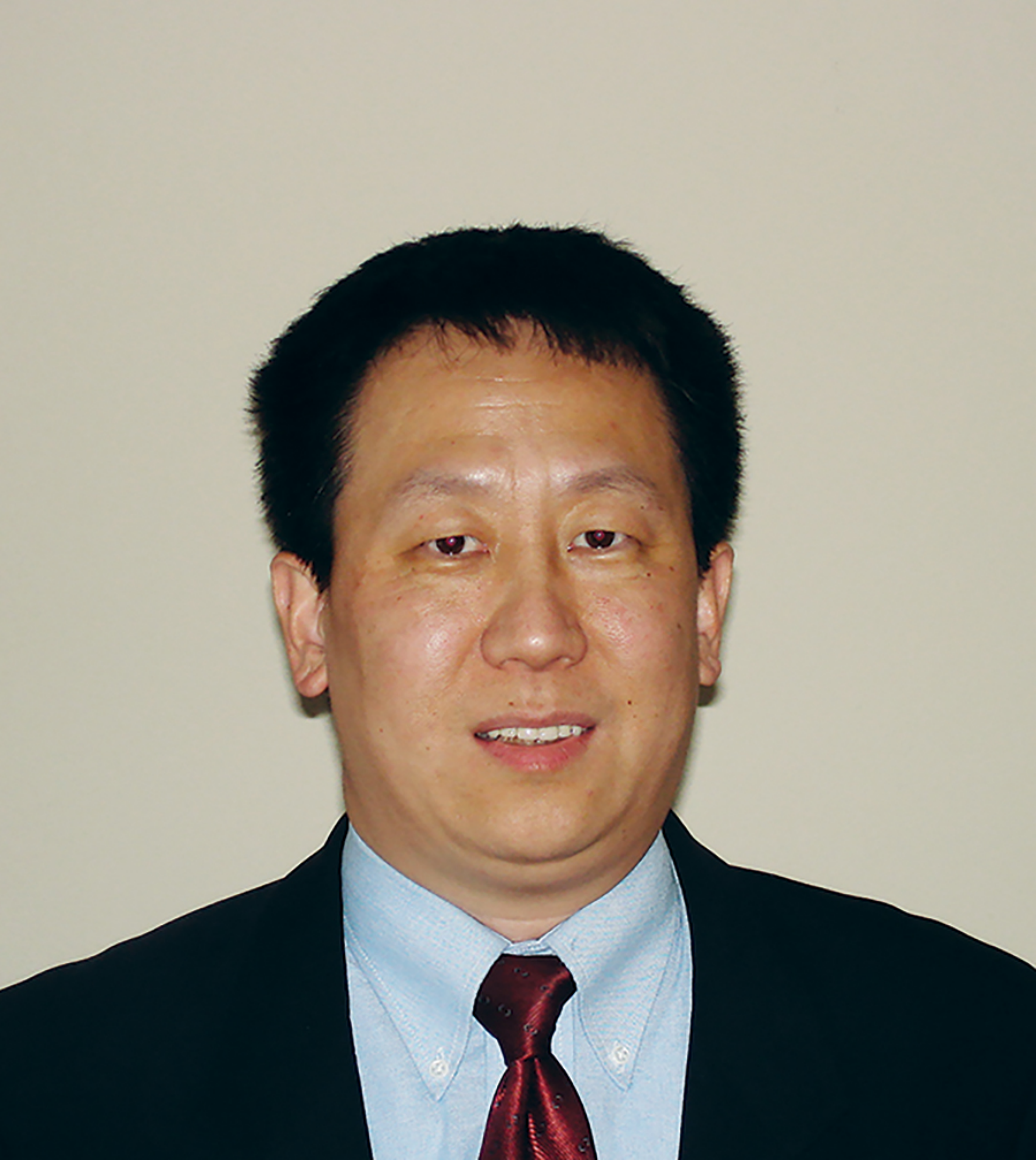}}]{Yu Ding}

(M'01, SM'11) received B.S. from University of Science \& Technology of China (1993); M.S. from Tsinghua University, China (1996); M.S. from Penn State University (1998); received Ph.D. in Mechanical Engineering from University of Michigan (2001). He is currently the Mike and Sugar Barnes Professor of Industrial \& Systems Engineering and a Professor of Electrical \& Computer Engineering at Texas A\&M University. His research interests are in data and quality science. Dr. Ding is the Editor-in-Chief of \emph{IISE Transactions} for the term of 2021--2024. Dr. Ding is a fellow of IIE, a fellow of ASME, a senior member of IEEE, and a member of INFORMS.

\end{IEEEbiography}

\begin{IEEEbiography}
	[{\includegraphics[width=1in,height=1.25in,clip,keepaspectratio]{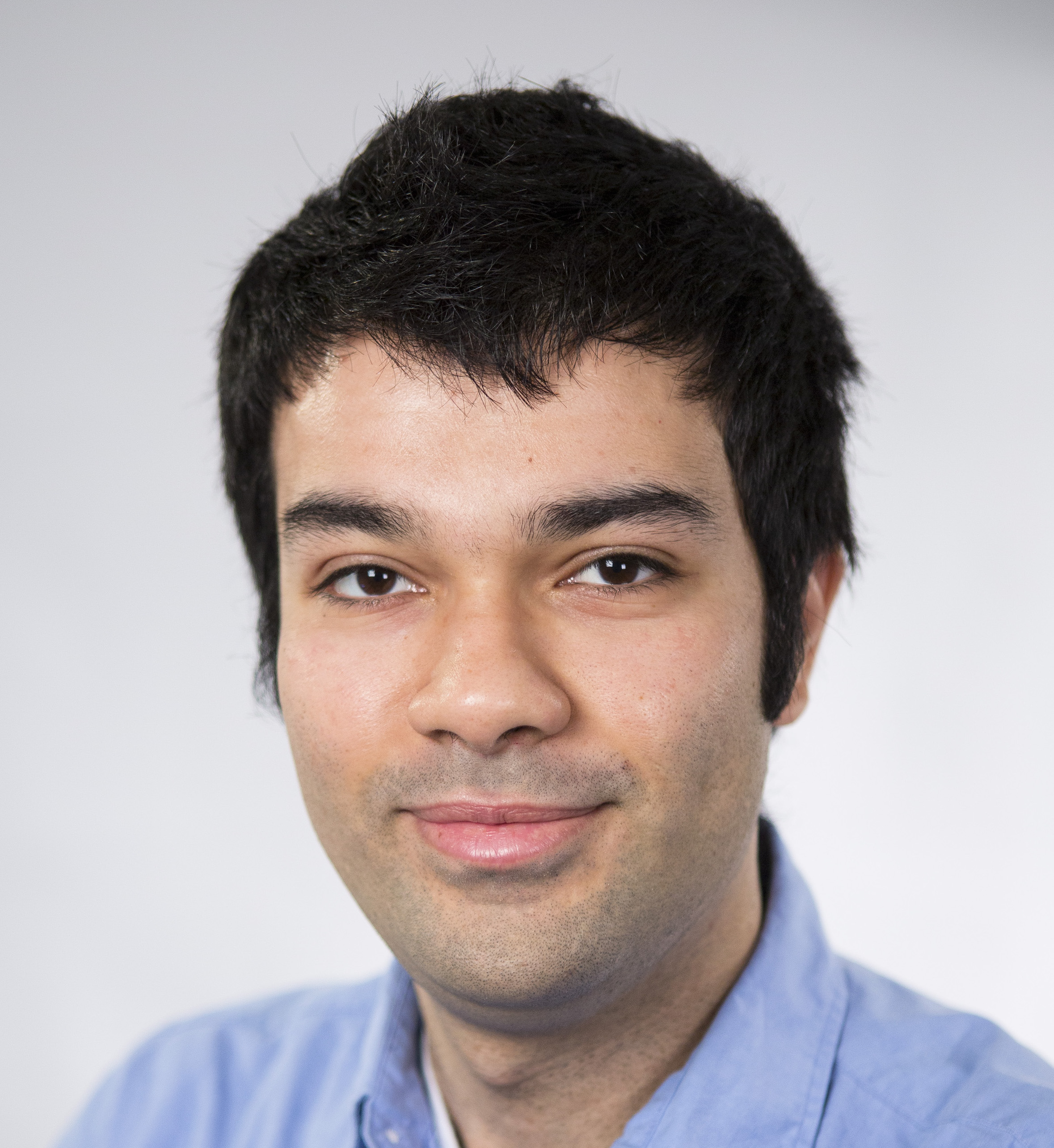}}] {Shahin Shahrampour} received the Ph.D. degree in Electrical and Systems Engineering, the M.A. degree in Statistics (The Wharton School), and the M.S.E. degree in Electrical Engineering, all from the University of Pennsylvania, in 2015, 2014, and 2012, respectively. He is currently an Assistant Professor in the Department of Mechanical and Industrial Engineering at Northeastern University. His research interests include machine learning, optimization, sequential decision-making, and distributed learning, with a focus on developing computationally efficient methods for data analytics. He is a Senior Member of the IEEE.
	\end{IEEEbiography}

\end{document}

%% file: header.tex
\newcommand\numberthis{\addtocounter{equation}{1}\tag{\theequation}}


\newcommand{\alg}{\text{ELSS}}


\newcommand{\0}{\mathbb{0}}
\newcommand{\1}{\mathbb{1}}

\newcommand{\E}{\mathbb{E}}
\newcommand{\R}{\mathbb{R}}
\renewcommand{\P}{\mathbb{P}}
\newcommand{\U}{\mathbb{U}}


\newcommand{\ab}{\mathbf{a}}
\newcommand{\bb}{\mathbf{b}}
\newcommand{\eb}{\mathbf{e}}
\newcommand{\ib}{\mathbf{i}}
\newcommand{\pb}{\mathbf{p}}
\newcommand{\qb}{\mathbf{q}}
\newcommand{\vb}{\mathbf{v}}
\newcommand{\ub}{\mathbf{u}}
\newcommand{\xb}{\mathbf{x}}
\newcommand{\yb}{\mathbf{y}}
\newcommand{\zb}{\mathbf{z}}

\newcommand{\Ab}{\mathbf{A}}
\newcommand{\Bb}{\mathbf{B}}
\newcommand{\Cb}{\mathbf{C}}
\newcommand{\Db}{\mathbf{D}}
\newcommand{\Eb}{\mathbf{E}}
\newcommand{\Fb}{\mathbf{F}}
\newcommand{\Gb}{\mathbf{G}}
\newcommand{\Hb}{\mathbf{H}}
\newcommand{\Ib}{\mathbf{I}}
\newcommand{\Jb}{\mathbf{J}}
\newcommand{\Kb}{\mathbf{K}}
\newcommand{\Lb}{\mathbf{L}}
\newcommand{\Pb}{\mathbf{P}}
\newcommand{\Qb}{\mathbf{Q}}
\newcommand{\Rb}{\mathbf{R}}
\newcommand{\Sb}{\mathbf{S}}
\newcommand{\Vb}{\mathbf{V}}
\newcommand{\Ub}{\mathbf{U}}
\newcommand{\Wb}{\mathbf{W}}
\newcommand{\Xb}{\mathbf{X}}
\newcommand{\Yb}{\mathbf{Y}}
\newcommand{\Zb}{\mathbf{Z}}

\newcommand{\alphab}{\boldsymbol{\alpha}}
\newcommand{\betab}{\boldsymbol{\beta}}
\newcommand{\gammab}{\boldsymbol{\gamma}}
\newcommand{\phib}{\boldsymbol{\phi}}
\newcommand{\Phib}{\boldsymbol{\Phi}}
\newcommand{\Qhib}{\boldsymbol{\Qhi}}
\newcommand{\omegab}{\boldsymbol{\omega}}
\newcommand{\psib}{\boldsymbol{\psi}}
\newcommand{\sigmab}{\boldsymbol{\sigma}}
\newcommand{\nub}{\boldsymbol{\nu}}
\newcommand{\thetab}{\boldsymbol{\theta}}
\newcommand{\delb}{\boldsymbol{\delta}}
\newcommand{\rhob}{\boldsymbol{\rho}}
\newcommand{\Pib}{\boldsymbol{\Pi}}
\newcommand{\pib}{\boldsymbol{\pi}}
\newcommand{\Sigmab}{\boldsymbol{\Sigma}}


\newcommand{\Bc}{\mathcal{B}}
\newcommand{\Cc}{\mathcal{C}}
\newcommand{\Dc}{\mathcal{D}}
\newcommand{\Ec}{\mathcal{E}}
\newcommand{\Fc}{\mathcal{F}}
\newcommand{\Hc}{\mathcal{H}}
\newcommand{\Lc}{\mathcal{L}}
\newcommand{\Nc}{\mathcal{N}}
\newcommand{\Oc}{\mathcal{O}}
\newcommand{\Pc}{\mathcal{P}}
\newcommand{\Rc}{\mathcal{R}}
\newcommand{\Sc}{\mathcal{S}}
\newcommand{\Tc}{\mathcal{T}}
\newcommand{\Uc}{\mathcal{U}}
\newcommand{\Xc}{\mathcal{X}}
\newcommand{\Yc}{\mathcal{Y}}


\newcommand{\tauh}{\widehat{\tau}}
\newcommand{\Sigmah}{\widehat{\Sigma}}

\newcommand{\fh}{\widehat{f}}
\newcommand{\gh}{\widehat{g}}
\newcommand{\kh}{\widehat{k}}
\newcommand{\qh}{\widehat{q}}
\newcommand{\Rh}{\widehat{R}}


\newcommand{\alphabh}{\widehat{\boldsymbol{\alpha}}}
\newcommand{\thetabh}{\widehat{\boldsymbol{\theta}}}

\newcommand{\qbh}{\widehat{\mathbf{q}}}

\newcommand{\Kbh}{\widehat{\mathbf{K}}}


\newcommand{\Fch}{\widehat{\mathcal{F}}}


\newcommand{\argmin}{\text{argmin}}
\newcommand{\arginf}{\text{arginf}}
\newcommand{\argmax}{\text{argmax}}
\newcommand{\minimize}{\text{minimize}}
\newcommand{\maximize}{\text{maximize}}
\newcommand{\supp}{\text{supp}}


\newcommand{\TV}{\text{TV}}
\newcommand{\norm}[1]{\left\lVert#1\right\rVert}
\newcommand{\tr}[1]{\text{Tr}\left[#1\right]}
\newcommand{\inn}[1]{\left<#1\right>}
\newcommand{\seal}[1]{\left \lceil #1\right \rceil}
\newcommand{\floor}[1]{\left \lfloor #1\right \rfloor}
\newcommand{\abs}[1]{\left|#1\right|}
\newcommand{\ind}[1]{\mathbf{1}\left(#1\right)}
\newcommand{\ex}[1]{\E\left[#1\right]}


\newtheorem{theorem}{Theorem}
\newtheorem{acknowledgement}[theorem]{Acknowledgement}
\newtheorem{assumption}{Assumption}
\newtheorem{conjecture}[theorem]{Conjecture}
\newtheorem{corollary}[theorem]{Corollary}
\newtheorem{definition}{Definition}
\newtheorem{example}{Example}
\newtheorem{lemma}[theorem]{Lemma}
\newtheorem{fact}{Fact}
\newtheorem{problem}{Problem}
\newtheorem{proposition}[theorem]{Proposition}
\newtheorem{remark}{Remark}
\newtheorem{solution}[theorem]{Solution}
\newtheorem{summary}[theorem]{Summary}

\newcommand{\bl}{\color{blue}}
\newcommand{\rd}{\color{black}}